\documentclass[sigconf,nonacm]{aamas} 
\usepackage{balance} 
\usepackage{hyperref}
\usepackage{url}

\usepackage{pifont}
\usepackage[utf8]{inputenc}
\usepackage[T1]{fontenc}
\usepackage{booktabs}
\usepackage{array}
\usepackage{multirow}
\usepackage{caption}
\usepackage{subcaption}
\usepackage{threeparttable}
\usepackage{siunitx}
\usepackage{longtable}
\usepackage{lscape}
\usepackage{graphicx}
\usepackage{caption}
\usepackage{verbatim}

\newcommand\blfootnote[1]{%
  \begingroup
  \renewcommand\thefootnote{}\footnote{#1}%
  \addtocounter{footnote}{-1}%
  \endgroup
}

\acmSubmissionID{1103}

\title{X-Ego: Acquiring Team-Level Tactical Situational Awareness via Cross-Egocentric Contrastive Video Representation Learning}

\author{Yunzhe Wang}
\affiliation{
  \institution{University of Southern California}
  \city{Los Angeles}
  \country{United States}}
\email{yunzhewa@usc.edu}

\author{Soham Hans}
\affiliation{
  \institution{University of Southern California}
  \city{Los Angeles}
  \country{United States}}
\email{sohamhan@usc.edu}

\author{Volkan Ustun}
\affiliation{
  \institution{USC Institute for Creative Technologies}
  \city{Los Angeles}
  \country{United States}}
\email{ustun@ict.usc.edu}

\begin{abstract}
Human team tactics emerge from each player's individual perspective and their ability to anticipate, interpret, and adapt to teammates's intentions. While advances in video understanding have improved the modeling of team interactions in sports, most existing work relies on third-person broadcast views and overlooks the synchronous, egocentric nature of multi-agent learning. We introduce \textbf{X-Ego-CS}, a benchmark dataset consisting of 124 hours of gameplay footage from 45 professional-level matches of the popular e-sports game \textit{Counter-Strike~2}, designed to facilitate research on multi-agent decision-making in complex 3D environments. X-Ego-CS provides \textbf{cross-egocentric} video streams that synchronously capture all players' first-person perspectives, along with state-action trajectories. Building on this resource, we propose \textbf{Cross-Ego Contrastive Learning (CECL)}, which aligns teammates' egocentric visual streams to foster team-level tactical situational awareness from an individual's perspective. We evaluate CECL on a teammate-opponent location prediction task, demonstrating its effectiveness in enhancing agent's ability to infer both teammate and opponent positions from a single first-person view on state-of-the-art video encoders. Together, X-Ego-CS and CECL establish a foundation for cross-egocentric multi-agent benchmarking in esports. More broadly, our work positions gameplay understanding as a testbed for multi-agent modeling and tactical learning, with implications for spatiotemporal reasoning and human-AI teaming in both virtual and real-world domains. Code and dataset are available at \url{https://github.com/HATS-ICT/x-ego}
\end{abstract}

\keywords{Contrastive Learning, Video Understanding, Gameplay Understanding, Teammate Modeling, Multi-Agent Systems, Esports Analytics}

\newcommand{\BibTeX}{\rm B\kern-.05em{\sc i\kern-.025em b}\kern-.08em\TeX}

\begin{document}


\pagestyle{fancy}
\fancyhead{}


\maketitle 






\section{Introduction}

\blfootnote{Preprint. Under review.}

\begin{figure}[h]
    \centering
    \vspace{7mm}
    \includegraphics[width=0.95\linewidth]{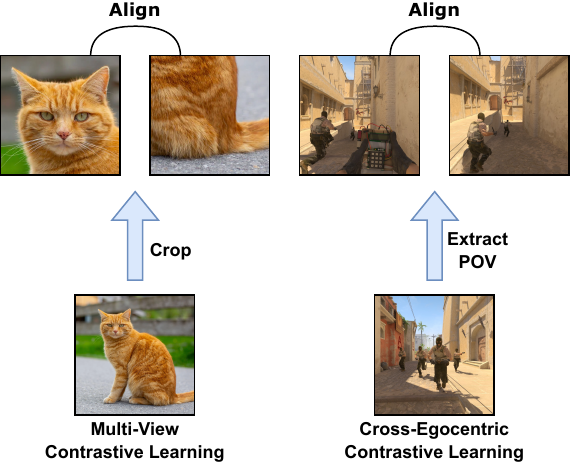}
    \caption{Illustration of Cross-Ego Contrastive Learning (CECL), where teammates' egocentric video representations are aligned across teams.}
    \label{fig:CECL}
\end{figure}

Team tactics are a defining feature of many human activities, from competitive sports to defense operations, emergency response, and cooperative games. Understanding and modeling such tactics requires reasoning not only about high-level joint actions and spatiotemporal coordination, but also about each agent's capacity to infer and anticipate the beliefs, intentions, and reactions of both teammates and opponents—a capability often referred to as Computational Theory of Mind \cite{rabinowitz2018machine, baker2011bayesian, jara2016naive, gurney2021operationalizing}. Early computational approaches to teamwork primarily focused on learning joint policies within simulated Multi-Agent Reinforcement Learning (MARL) frameworks \cite{lowe2017multi,leibo2017multi,yang2018mean,rashid2020monotonic,vinyals2019grandmaster,berner2019dota}. However, these methods face two major limitations: (1) many simulation environments are grid-based or two-dimensional, simplifying real-world complexity; even in large-scale domains such as \textit{StarCraft II} \cite{vinyals2019grandmaster} and \textit{Dota 2} \cite{berner2019dota}, the worlds are presented through isometric or ``pseudo-3D'' perspectives rather than fully three-dimensional environments; (2) learning architectures seldom include explicit models of other agents' internal states. Consequently, while such systems can coordinate actions, they lack the ability to interpret or anticipate teammates' and opponents' mental states. This gap limits their scalability and transferability to real-world scenarios, which is essential for advancing the computational modeling of human team tactics.

In recent years, there has been growing interest in extending team-level behavior modeling to real-world competitive sports domains, particularly in soccer and basketball \cite{wang2024tacticai, deliege2021soccernet, rao2024matchtime, gu2020fine, decroos2018automatic} for tasks such as action recognition, event detection, tactical planning and discovery, and automated commentary generation. This trend is driven both by the complexity of these sports and by the economic and analytical value of understanding team strategies and player interactions. However, existing datasets and models still fall short of enabling true multi-agent understanding since most datasets rely on third-person broadcast footage from fixed angles with global visibility, which limits the study of individual perception, uncertainty, and coordination from a first-person perspective.

In contrast, First-Person Shooter (FPS) competitive video games such as \textit{Counter-Strike} provide a natural testbed and an effective middle ground that balances game-state richness and decision-making complexity, while offering reliable ground-truth data for studying team behavior under partial observability. The game provides a cooperative yet adversarial environment on fixed 3D maps, where players must continuously balance tactical risk, spatiotemporal dynamics, and coordination. While relatively underexplored, prior work has begun to investigate this space: \cite{xenopoulos2022esta} introduced the ESTA dataset along with the AWPY parser for extracting ground-truth state-action trajectories and game events, \cite{wang2025csgo} developed the DECOY simulator to support multi-agent reinforcement learning, \cite{durst2024learning} trained a transformer-based control model to mimic professional player movement, and \cite{fanourakis2025amucs} studies players' emotional dynamics with synchronized physiological, behavioral, and gameplay signals. While these efforts highlight the promise of the domain, they largely focus on individual agents or global game states, lacking a systematic approach to modeling the synchronous, egocentric nature of multi-agent team interactions.

To address this gap, we introduce \textbf{X-Ego-CS}, a benchmark dataset comprising 124 hours of gameplay footage from 45 professional-level \textit{Counter-Strike~2} matches, containing synchronized egocentric video streams from all players. Unlike prior datasets focused on single perspectives or third-person views, X-Ego-CS captures simultaneous first-person recordings paired with state-action trajectories sampled at 64 ticks per second. We define this multi-perspective video paradigm as \textbf{cross-egocentric}, combining the egocentric focus of individual perception with cross-view synchronization across teammates.

Building upon X-Ego-CS, we propose \textbf{Cross-Ego Contrastive Learning (CECL)}, a method designed to learn shared world-state representations by aligning teammates' egocentric visual streams at corresponding timesteps. This alignment encourages models to internalize a collective understanding of team context, which is an essential capability for agents aiming to exhibit \textit{theory-of-mind}-like reasoning and collaborative tactical awareness. We evaluate CECL on a downstream \textbf{teammate-opponent location prediction} task, demonstrating its effectiveness in enhancing agents' ability to infer both teammate and opponent positions from a single first-person view when applied to state-of-the-art video encoders. These results highlight CECL's potential to foster team-level situational awareness and establish \textbf{cross-egocentric multi-agent video understanding} as a promising direction for advancing tactical reasoning and human-AI teaming. Our contributions are twofold: 

\begin{itemize}
    \item We introduce X-Ego-CS, the first benchmark dataset for cross-egocentric multi-agent video understanding in esports, and established teammate-opponent location prediction task for teammate and opponent modeling.
    \item We propose CECL, a visual contrastive learning method that learns shared team-state representations by aligning entire teams' first-person views at the same timestep, boosting individual agents' team-level situational awareness.
\end{itemize}

\section{Related Work}

\paragraph{Sports Understanding} 

Computational analysis of team sports has traditionally focused on third-person broadcast views, particularly in soccer and basketball, addressing tasks such as action recognition \cite{giancola2018soccernet, deliege2021soccernet}, tactical planning \cite{wang2024tacticai}, and automated commentary \cite{mkhallati2023soccernet, rao2024matchtime, rao2025towards}. While these approaches have advanced strategic analysis and performance analytics, they rely on omniscient overhead perspectives that do not capture the egocentric, partially observable nature of real-time decision-making that players experience.

\paragraph{Gameplay Understanding} 

Research in gameplay understanding within AI can be broadly organized into two categories: (1) enabling AI agents to play games autonomously, and (2) facilitating effective collaboration between humans and AI. Significant progress in autonomous gameplay has been achieved through deep reinforcement learning in single-agent settings, from early successes in Atari \cite{mnih2015human} to mastering board games \cite{silver2016mastering, silver2017mastering}, as well as through multi-agent reinforcement learning in cooperative-competitive environments, including multiplayer online battle arena (MOBA) games \cite{vinyals2019grandmaster, berner2019dota}, soccer \cite{kurach2020google}, and emergent multi-agent behaviors \cite{baker2019emergent}. More recent advances have explored LLM-enabled gameplay \cite{wang2023voyager} and world modeling approaches \cite{hafner2023mastering, hafner2025training, bruce2024genie}. 

In parallel, human-AI collaboration has emerged as a critical area in games and robotics, with works exploring coordination challenges \cite{carroll2019utility}, human expectations of AI collaborators \cite{zhang2021ideal}, dual-process architectures for real-time teaming \cite{zhang2025leveraging}, shared mental model alignment tools \cite{gu2025enabling}, and multimodal research platforms \cite{zhang2024crew}. Recent work has also proposed interpretable behavioral task-space frameworks to evaluate human-AI alignment in large-scale multiplayer games \cite{sharma2024toward}. Analysis of teammate preferences reveals that interpretability can be as important as performance, with humans often favoring rule-based agents over learned ones despite comparable outcomes \cite{siu2021evaluation}.

Despite these advances, open challenges remain in scaling gameplay understanding to competitive esports and AAA-level games, particularly in reasoning over rich visual environments, fast-paced dynamics, and high degrees of freedom, and in transferring knowledge learned from gameplay to real-world human-AI teaming.

\paragraph{Contrastive Learning}

Contrastive learning has emerged as a powerful paradigm for self-supervised representation learning. Its core idea—learning by distinguishing positive from negative pairs—can be traced back to early energy-based models and noise-contrastive estimation \cite{lecun2006tutorial,gutmann2010noise}. In natural language processing, similar principles underpin word embedding methods such as Word2Vec \cite{mikolov2013efficient}, which train models to discriminate true word-context pairs from random ones. More recently, contrastive learning has achieved remarkable success in computer vision through frameworks like CPC \cite{oord2018representation}, SimCLR \cite{chen2020simple}, and MoCo \cite{he2020momentum}, which leverage data augmentations and large negative sets to learn transferable visual representations. Building on these advances, contrastive learning has also proven effective in multimodal settings, where it enables alignment between different modalities such as vision and language \cite{radford2021learning, zhai2023sigmoid}.

In the multi-agent learning domain, contrastive learning has gained increasing traction in recent years for addressing key challenges such as coordination, diversity, opponent modeling, and communication. Recent work explores how contrastive objectives can disentangle agent roles and enhance cooperation by promoting behavioral heterogeneity and more effective credit assignment \cite{hu2023attention}. Other studies investigate how contrastive trajectory representations can encourage diversity among agents that share parameters, leading to richer exploration and more robust team strategies \cite{li2024learning}. To improve adaptability in mixed cooperative-competitive settings, contrastive formulations have been applied to learn policy embeddings of teammates and opponents directly from single-agent observations, enabling faster response and generalization across tasks \cite{ma2025contrastive}. In addition, contrastive principles have been used to learn more effective communication protocols between agents, capturing shared environmental information and promoting symmetric message exchange in decentralized systems \cite{lo2023learning,yu2025tactic}. However, these approaches have primarily been validated in simplified reinforcement learning environments, raising important questions about their applicability and scalability in more realistic, high-complexity multi-agent scenarios.

\section{X-Ego-CS Dataset}

Our dataset comprises highly curated gameplay recordings including cross-egocentric video streams and structured state-action trajectories, all extracted from in-game replay demo files. We describe the data curation process in Section \ref{sec:data_curation}, and provide summary statistics in Section \ref{sec:statistics}. To our best knowledge, X-Ego-CS is the first dataset that contains synchronized egocentric video streams and structured state-action trajectories from all players in professional e-sports matches. A comparison with similar datasets is shown in Table \ref{tab:datasets}.

\begin{table}[ht]
  \centering
  \caption{Characteristic comparison of X-Ego-CS with similar video datasets. Symbols: \ding{51} = available, \ding{55} = not available, $\triangle$ = partially satisfied.}
  \label{tab:datasets}
  \footnotesize
  \setlength{\tabcolsep}{2pt}
  \renewcommand{\arraystretch}{1.2}
  \begin{tabular}{p{2.0cm}p{1.8cm}cccccc}
  \toprule
  \textbf{Dataset} & \textbf{Domain} & \textbf{Team} & \textbf{Expert} & \textbf{Video} & \textbf{Traj.} & \textbf{Ego-Cen.} \\
  \midrule
  ESTA \citep{xenopoulos2022esta} & ESports (CS:GO) & \ding{51} & \ding{51} & \ding{55} & \ding{51} & \ding{55} \\
  AMuCS \citep{fanourakis2025amucs} & ESports (CS:GO) & $\triangle$ & \ding{55} & \ding{51} & \ding{51} & \ding{51} \\
  SoccerNet \citep{giancola2018soccernet} & Sports (Soccer) & \ding{51} & \ding{51} & \ding{51} & $\triangle$ & \ding{55} \\
  SoccerReplay \citep{rao2025towards} & Sports (Soccer) & \ding{51} & \ding{51} & \ding{51} & \ding{55} & \ding{55} \\
  Waymo Open \citep{sun2020scalability} & Self-Driving Cars & \ding{55} & \ding{55} & \ding{51} & \ding{51} & \ding{51} \\
  Ego4D \citep{grauman2022ego4d} & Daily Activities & \ding{55} & \ding{55} & \ding{51} & \ding{51} & \ding{51} \\
  \midrule
  \textbf{X-Ego-CS (Ours)} & ESports (CS2) & \ding{51} & \ding{51} & \ding{51} & \ding{51} & \ding{51} \\
  \bottomrule
  \end{tabular}
\end{table}

\subsection{Automated Data Collection and Curation}
\label{sec:data_curation}

We collected professional-level Counter-Strike~2 data from in-game demo files (\texttt{.dem}), all downloaded from the popular CS2 matchmaking platform FACEIT \cite{faceit2025}. Using FACEIT's official API and following website policies, we retrieved the public Elo rating leaderboard and match history for the top 100 players, collecting their most recent matches. From these replays, we extracted structured metadata, player state-action trajectories, and game events using off-the-shelf parsers including AWPY and demoparser2 \cite{xenopoulos2022esta,LaihoE_demoparser_2025}. 

To ensure reproducible and standardized video data, we developed an automated in-game recording system built on top of the Counter-Strike~2 demo playback engine. The demo metadata provides exact tick ranges for every player's alive period within a round. By issuing console commands such as \texttt{demo\_gototick} to jump to the start tick, \texttt{spec\_player} to lock the camera onto the correct player, and \texttt{demo\_resume}/\texttt{demo\_pause} to control playback, we were able to deterministically replay only the segments corresponding to each player's active life. During playback, NVIDIA Shadowplay \cite{nvidia_app} was automatically triggered via hotkeys to record the alive duration of each player per round.

After filtering for quality, the final dataset contains 45 matches, all played on the \texttt{de\_mirage} map, including synchronized egocentric video recordings from all players and state-action trajectories.

\subsection{Statistics}
\label{sec:statistics}
\begin{figure}[!ht]
    \centering
    \includegraphics[width=0.99\linewidth]{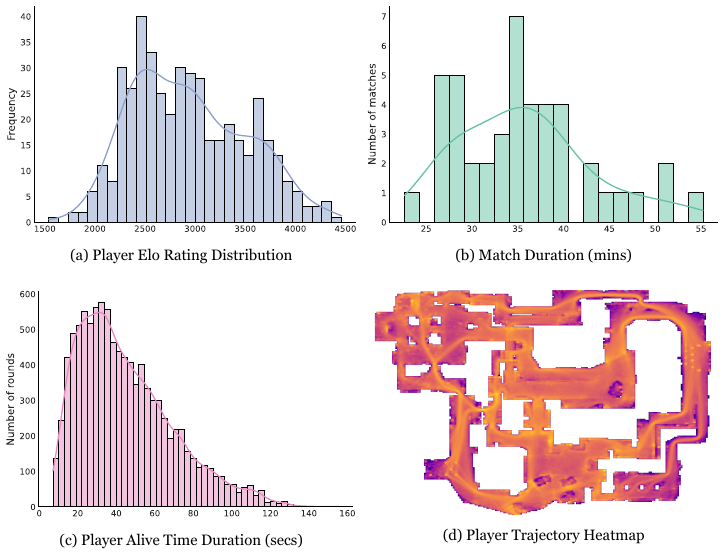}
    \caption{Additional statistics of the X-Ego-CS dataset. (a) Player FACEIT Elo Rating Distribution: above 2000 corresponds to the highest FACEIT level 10 rank (roughly top 10\% of players), and above 3000 represents roughly the top 1\%. (b) Match Duration. (c) Player Alive Time Duration, which corresponds to the duration of video files. (d) Player Trajectory Heatmap aggregated from both team sides across all matches, visualized on the de\_mirage map.}
    \label{fig:stats}
\end{figure}

X-Ego-CS contains 45 curated professional-level matches spanning 1011 rounds and approximately 124 hours of gameplay video footage. Each match includes cross-egocentric video streams, structured state-action trajectories, and round-level event annotations, covering 372 unique players, with an average round duration of 44 seconds of player alive time. All videos are recorded at 720p and 30 FPS. Additional statistics are shown in Figure \ref{fig:stats}.

\section{Method}

In this paper, we aim to learn a visual representation that enables individual agents to achieve team-level situational awareness from their egocentric observations. We first formulate cross-egocentric tactical understanding as a team-based representation learning task (Sec.~\ref{sec:problem_formulation}), then describe our model architecture (Sec.~\ref{sec:model_architecture}) and Cross-Egocentric Contrastive Learning objective (Sec.~\ref{sec:cross_egocentric_contrastive_learning}), and finally present the downstream teammate-opponent location prediction tasks (Sec.~\ref{sec:teammate_opponent_location_prediction}).

\subsection{Problem Formulation}
\label{sec:problem_formulation}
We address the problem of multi-agent video understanding and teammate modeling with RGB video segments $\mathcal{V} \in \mathbb{R}^{A \times T \times 3 \times H \times W}$ from each team, where $A$ denoting the number of agents and $T$ denoting the number of frames. Our goal is to learn a visual encoder $\Phi_\text{vision}$ that maps each agent's egocentric video segment to an embedding space $\mathbf{Z}$, such that $\text{Sim}(Z_i, Z_j)$ is maximized for teammates $i,j \in \mathcal{T}$ at the same time segment and minimized for agents otherwise. After alignment, we can take a single agent's embedding $Z_i$ or a subset of agents from the same team to infer information about all teammates and opponents. We hypothesize that the contrastive objective encourages implicit information sharing among teammates by aligning their egocentric representations toward a shared latent team state. This process allows the encoder to capture common situational cues such as formation, timing, tactical phase.

Consider a flashbang grenade event that causes synchronous blindness among agents in the flash's line of sight. When multiple teammates are simultaneously flashed, their first-person visual streams exhibit similar characteristics (whiteout effects and reduced motion). The contrastive objective maximizes $Sim(Z_i, Z_j)$ for these affected teammates, which forces $\Phi_\text{vision}$ to produce similar embeddings for the shared visual patterns. Critically, the encoder cannot simply memorize individual whiteout frames—it must learn that such synchronized sensory disruptions encode tactical information: the grenade's origin and trajectory constrain opponent positions, and coordinated flashes typically indicate clustered team movements during site pushes. CECL trains the encoder to map these shared visual patterns to a latent representation of the underlying team state. Even when only one player's view is available during inference, the learned embedding space could help the model better infer teammate proximity and opponent positioning, as the alignment process and downsteam objectives encodes relevant spatial configurations correlated with such sensory patterns.

\subsection{Model Architecture}
\label{sec:model_architecture}

\begin{figure*}[ht]
  \centering
  \includegraphics[width=0.99\linewidth]{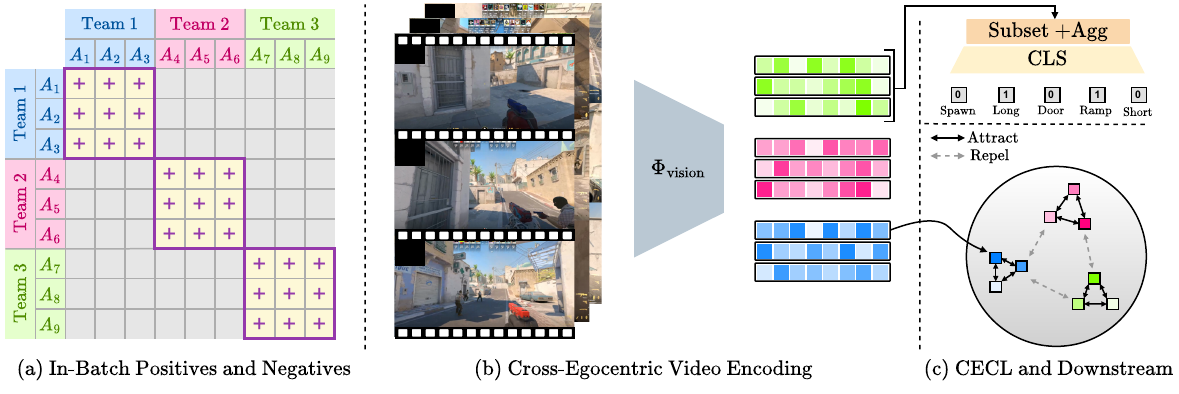}
  \caption{Illustration of the forward pass on a data batch containing three teams of three agents each. (a) Agents from the same team are treated as positive pairs, while agents from different teams or time segments are treated as negatives. (b) Each agent's egocentric video segment is encoded into a representation vector; a mask is applied to the top-left corner to prevent leaking teammate or opponent locations from the mini-map. (c) The CECL objective aligns teammates' egocentric video representations within the same batch while repelling others. A subset of agents from a team is then selected for team-level location prediction, formulated as a multi-label classification task.}
  \label{fig:x-ego}
\end{figure*}

Our model is built upon state-of-the-art vision transformer architecture (Sec \ref{sec:Implementation}) that processes egocentric video streams. The overall architecture consists of three main components: (1) a spatiotemporal video encoder that extracts frame-level and temporal features from each player's first-person view, (2) a contrastive projection head that maps these features into a shared embedding space where team-based alignment is enforced, and (3) downstream MLP predictor heads that predict either teammate or opponent locations from subset and aggregated agent embeddings (Fig~\ref{fig:x-ego}).

A visual encoder-projector network $\Phi_\text{vision}$ extracts spatiotemporal representations for each agent, producing embeddings 
\[
E = \Phi(V) \in \mathbb{R}^{A \times d_{\text{proj}}}
\]

During inference, a subset of \(N\) agents's visual embeddings is selected, where \(N \leq A\), and for the single-agent case \(N = 1\). When \(N > 1\), the selected embeddings are aggregated through an agent aggregator $\phi_{\text{agg}}$ with is concatenation plus two-layer MLP. 
The resulting representation is concatenated with the team-side embedding \(S \in \mathbb{R}^{d_s}\) to form the combined feature
\[
Z = [F; S]
\]
where \(S\) is a single vector indicating the team side of the POV (either T team or CT team). This combined feature is then processed by task-specific heads 
\[
\Psi = \{\Psi_{\text{TLN}}, \Psi_{\text{ELN}}\}
\]
corresponding to the \textbf{Teammate Location Nowcast (TLN)} and \textbf{Enemy Location Nowcast (ELN)} tasks. This formulation allows the unified model \(\Psi(\Phi(V))\) to flexibly adapt to single-agent or multi-agent subsets for downstream objectives.

\subsection{Cross-Egocentric Contrastive Learning}  
\label{sec:cross_egocentric_contrastive_learning}

We adopt a sigmoid-based contrastive objective \cite{zhai2023sigmoid} to align cross-egocentric representations within teams. Given a video encoder $f(\cdot)$, for each player $i$ in a batch, we obtain L2-normalized embeddings $\mathbf{u}_i = \tfrac{f(V_i)}{\|f(V_i)\|_2}$ from their egocentric video segment. We employ a \textit{multi-positive contrastive strategy}, where all teammates observing the same round at the same time segment serve as positive pairs (Fig~\ref{fig:x-ego}a). This naturally creates a three-level hierarchy of negatives: (1) same team and round but different time segments, (2) different team (i.e., opponents) at the same or different time, and (3) different rounds or matches. The contrastive objective is:

\begin{equation}
  \label{eq:cecl}
\mathcal{L}_{\text{CECL}}
=
-\frac{1}{|\mathcal{B}|}
\sum_{i=1}^{|\mathcal{B}|}
\sum_{j=1}^{|\mathcal{B}|}
\log\frac{1}{1+e^{m_{ij}(-t\,\mathbf{u}_i\cdot \mathbf{u}_j + b)}}
\end{equation}

Here, $\mathbf{u}_i \cdot \mathbf{u}_j$ represents the cosine similarity between player $i$ and player $j$'s egocentric embeddings. The parameter $t$ is a learnable temperature for scaling, and $b$ is a learnable bias. The label $m_{ij}$ equals $1$ if players $i$ and $j$ are on the same team observing the same round at the same time segment (positives), and $-1$ otherwise (negatives). We initialize $b$ to $-3$ and $t$ to $\log(10)$. This initialization accounts for the large imbalance between positive and negative pairs where negatives dominate the loss, with a positive ratio of $p_+ = |\mathcal{B}|/|\mathcal{B}|^2 = 1/|\mathcal{B}|$. We initialize the bias to roughly the prior log-odds of the positive ratio, i.e., $\text{logit}(p_+) = \log(p_+/(1-p_+))$, which in our case is roughly $-3$. We also conducted an ablation of the bias terms in Section \ref{sec:ablation} that confirms the design choice.

\subsection{Teammate-Opponent Location Prediction}
\label{sec:teammate_opponent_location_prediction}

We designed a video-based teammate-opponent location prediction task to evaluate the effectiveness of CECL. Given an agent's egocentric video segment, Teammate Location Nowcast (TLN) aims to predict the spatial positions of all teammates at the corresponding time step, while Opponent Location Nowcast (OLN) aims to infer the positions of all adversarial agents.

Formally, let $\mathcal{L} = \{l_1, l_2, \ldots, l_N\}$ denote the set of $N$ discrete location identifiers in the environment. At time step $t$, given an agent's egocentric video segment $\mathbf{v}_t$, we formulate the location nowcasting task as a multi-label classification problem. For each location $l_i \in \mathcal{L}$, we predict a binary occupation label $y_i \in \{0, 1\}$, where $y_i = 1$ indicates that at least one agent (teammate for TLN, opponent for OLN) currently occupies location $l_i$ at time $t$, and $y_i = 0$ otherwise. The model outputs a prediction vector $\mathbf{y}_t = [y_1, y_2, \ldots, y_N]^T \in \{0,1\}^N$, representing the concurrent occupancy distribution across all locations. In X-Ego-CS, on the de\_mirage map, there are $N=23$ locations.

\paragraph{Classification Labels}

We acquire location labels from the game trajectory data, where each 3D coordinate has a corresponding categorical area identifier. The \texttt{de\_mirage} map contains 23 distinct locations with names corresponding to strategic callouts that players commonly use during gameplay, such as \texttt{T\_Spawn}, \texttt{CT\_Spawn}, \texttt{Bombsite\_A}, \texttt{Bombsite\_B}, \texttt{Catwalk}, \texttt{Stairs}, \texttt{Connector}, etc. To construct ground truth labels, we extract each agent's 3D coordinates at time $t$ and map them to their categorical area name. When an agent's trajectory spans multiple areas within a video segment, we select the area at the middle temporal point to ensure alignment. For simplicity, we only use video segments where all 10 players are alive. The final ground truth $\mathbf{y}_t \in \{0,1\}^{23}$ is a binary vector where each dimension indicates whether at least one agent occupies the corresponding location.

\paragraph{Evaluation Metrics}
We report standard metrics for multi-label classification: Subset Accuracy, which measures the fraction of samples where all location predictions exactly match the ground truth labels across all 23 locations; Hamming Distance, which quantifies the fraction of misclassified labels across all location predictions, and Micro and Macro F1 scores, averaged over all labels.

\section{Experiments}

\begin{table*}[!ht]
  \centering
  \caption{Performance comparison of CECL vs. baseline models on Teammate Location Nowcast and Enemy Location Nowcast tasks. Bold values indicate best results; red values highlight improvements over baseline.}
  \label{tab:phase1-main}
  \small
  \begin{tabular}{lcccccccc}
  \toprule
  \multirow{2}{*}{\textbf{Method}} & \multicolumn{4}{c}{\textbf{Teammate Location Nowcast}} & \multicolumn{4}{c}{\textbf{Enemy Location Nowcast}} \\
  \cmidrule(lr){2-5}\cmidrule(lr){6-9}
  & \textbf{Sub.Acc} $\uparrow$ & \textbf{Ham.Dist} $\downarrow$ & \textbf{Macro F1} $\uparrow$ & \textbf{Micro F1} $\uparrow$ & \textbf{Sub.Acc} $\uparrow$ & \textbf{Ham.Dist} $\downarrow$ & \textbf{Macro F1} $\uparrow$ & \textbf{Micro F1} $\uparrow$ \\
  \midrule
  \multicolumn{9}{c}{Off-the-shelf Models (Frozen) + Linear Probe} \\
  \midrule
  DINOv2 \citep{oquab2023dinov2}               & 16.89 & 9.90 & 35.66 & 57.19 & 13.47 & \textbf{11.13} & \textbf{22.92} & \textbf{49.77} \\
  ViViT \citep{arnab2021vivit}                 & 16.97 & 10.33 & 35.69 & 56.12 & \textbf{14.21} & 11.30 & 22.83 & 48.14 \\
  VideoMAE \citep{tong2022videomae}              & 12.92 & 11.46 & 21.00 & 47.25 & 12.57 & 11.50 & 18.10 & 45.64 \\
  SigLIP \citep{zhai2023sigmoid}                & \textbf{17.67} & \textbf{9.58} & \textbf{39.77} & \textbf{59.40} & 11.02 & 11.37 & 19.90 & 46.91 \\
  \midrule
  \multicolumn{9}{c}{Cross-Ego Contrastive Learning} \\
  \midrule
  CECL (DINOv2)  & \textbf{18.13} \textcolor{red}{(+1.24)} & \textbf{9.67} \textcolor{red}{(-0.23)} & \textbf{37.61} \textcolor{red}{(+1.95)} & \textbf{58.37} \textcolor{red}{(+1.18)} & \textbf{14.60} \textcolor{red}{(+1.13)} & \textbf{11.00} \textcolor{red}{(-0.13)} & \textbf{25.02} \textcolor{red}{(+2.10)} & \textbf{50.54} \textcolor{red}{(+0.77)} \\
  CECL (ViViT)   & 18.05 \textcolor{red}{(+1.08)} & 10.06 \textcolor{red}{(-0.27)} & 37.31 \textcolor{red}{(+1.62)} & 57.26 \textcolor{red}{(+1.15)} & 12.92 (-1.28) & 11.39 (+0.09) & 20.43 (-2.41) & 47.72 (-0.41) \\
  CECL (VideoMAE) & 14.29 \textcolor{red}{(+1.36)} & 11.18 \textcolor{red}{(-0.29)} & 22.08 \textcolor{red}{(+1.08)} & 47.26 \textcolor{red}{(+0.01)} & 13.04 \textcolor{red}{(+0.47)} & 11.38 \textcolor{red}{(-0.12)} & 20.18 \textcolor{red}{(+2.08)} & 47.09 \textcolor{red}{(+1.45)} \\
  CECL (SigLIP)  & 17.94 \textcolor{red}{(+0.27)} & 10.00 (+0.43) & 35.88 (-3.90) & 57.11 (-2.29) & 13.47 \textcolor{red}{(+2.45)} & 11.14 \textcolor{red}{(-0.23)} & 24.49 \textcolor{red}{(+4.59)} & 49.64 \textcolor{red}{(+2.72)} \\
  \bottomrule
  \end{tabular}
\end{table*}

\begin{figure*}[ht]
  \centering
  \includegraphics[width=0.95\linewidth]{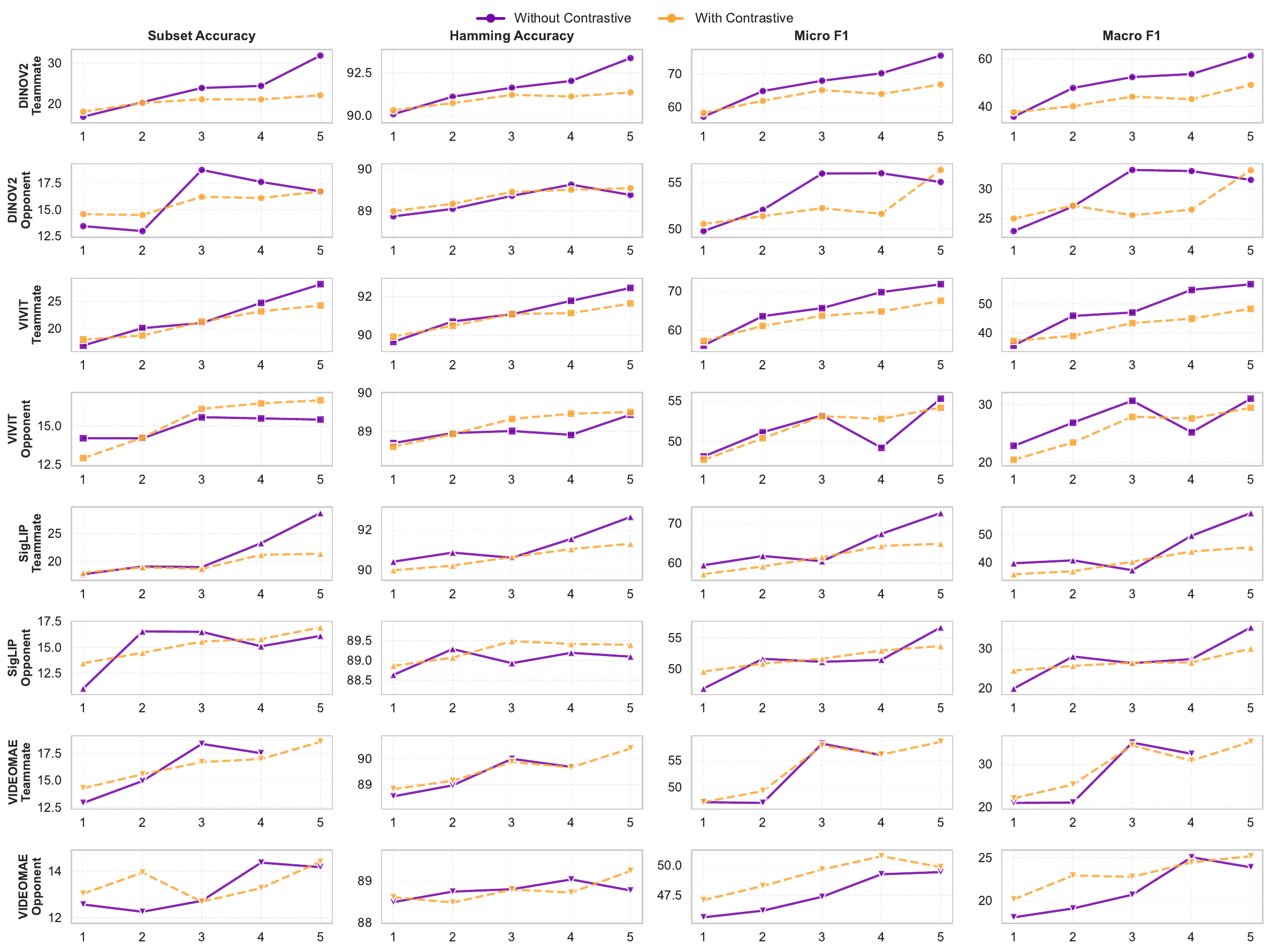}
  \caption{Performance comparison of CECL vs. baseline across different visual encoders (DINOv2, ViViT, VideoMAE, SigLIP) as the number of agent POVs varies from 1 to full team size of 5. CECL shows substantial gains at lower POV counts (1-2 agents) but diminishing returns as team coverage increases.}
  \label{fig:multi-agent}
\end{figure*}

We evaluate CECL across multiple dimensions. We first describe the implementation details including encoder architectures, data processing, and training configurations (Sec.~\ref{sec:Implementation}). We then present quantitative results comparing CECL against baseline models on single-POV and multi-POV settings (Sec.~\ref{sec:quantitative}), followed by ablation studies on key hyperparameters (Sec.~\ref{sec:ablation}). Finally, we provide qualitative analysis through embedding space visualizations (Sec.~\ref{sec:qualitative}).

\subsection{Implementation Details}
\label{sec:Implementation}

\paragraph{Video Encoders}
We evaluate our cross-egocentric contrastive learning framework using four state-of-the-art video encoders: SigLIP \citep{zhai2023sigmoid}, DINOv2 \citep{oquab2023dinov2}, ViViT \citep{arnab2021vivit}, VideoMAE \citep{tong2022videomae}. For image-based models (SigLIP, DINOv2), we compute temporal embeddings by mean pooling frame-level representations. For video models with fixed temporal receptive fields (ViViT, VideoMAE), we apply temporal resampling to match the target sequence length through interpolation or truncation.

All video encoders are kept frozen during training, and we learn a two-layer MLP projector $\phi_{\text{proj}}: \mathbb{R}^{d_{\text{enc}}} \rightarrow \mathbb{R}^{768}$ for contrastive alignment.

\paragraph{Data Processing}
Videos are preprocessed through a standardized pipeline: (1) we resize videos to $224 \times 224$ resolution with bilinear interpolation, distorting the aspect ratio; (2) pixel values are normalized using ImageNet statistics ($\mu = [0.485, 0.456, 0.406]$, $\sigma = [0.229, 0.224, 0.225]$); and (3) temporal sampling at 4 FPS over 5-second windows with $\pm 0.3$ second jittering for temporal robustness. The dataset is partitioned with a 70:15:15 train:validation:test split based on total rounds to ensure temporal independence. We mask out the top-left corner of all videos to ensure that teammate and opponent locations are not leaked in the mini-map.

\paragraph{Training Configuration}
We employ the AdamW optimizer \citep{loshchilov2017decoupled} with learning rate $\eta = 3 \times 10^{-4}$, weight decay $\lambda = 10^{-2}$, and bFloat16 data type with mixed-precision training. We train all models for 8 epochs and select the best checkpoint based on validation performance. We use batch size $B = 32$ and apply gradient accumulation when memory constraints require smaller effective batch sizes. All experiments are conducted on single GPUs including A100, A40, or RTX4090.

For the multi-label classification tasks Teammate Location Nowcast and Enemy Location Nowcast, we optimize the Binary Cross Entropy loss:
\[
\mathcal{L}_{\text{BCE}} = -\frac{1}{N} \sum_{i=1}^{N} \sum_{j=1}^{L} [y_{ij} \log(\hat{p}_{ij}) + (1-y_{ij}) \log(1-\hat{p}_{ij})]
\]
where $N$ is the batch size, $L = 23$ is the number of location classes, $y_{ij} \in \{0,1\}$ are ground truth labels, and $\hat{p}_{ij} \in [0,1]$ are predicted probabilities. The total loss combining with CECL (eq.~\ref{eq:cecl}) is:
\[
\mathcal{L}_{\text{total}} = \lambda \cdot \mathcal{L}_{\text{CECL}} + \mathcal{L}_{\text{BCE}}
\]
where $\lambda$ is a balancing weighting coefficient.

\subsection{Quantitative Evaluation}
\label{sec:quantitative}

\paragraph{Single POV Evaluation}

Table~\ref{tab:phase1-main} shows the performance comparison of CECL vs. off-the-shelf state-of-the-art visual models on Teammate Location Nowcast and Enemy Location Nowcast tasks across 4 multi-label classification metrics, using only single-agent POV visual embeddings as input. The results demonstrate that the CECL learning objective generally boosts performance at the single-agent POV level across most model and metric combinations, with a mean absolute gain of 1.45 percentage points over all baselines.

\paragraph{Multi-POV Evaluation}

We further evaluate the effectiveness of CECL by varying the number of agent POVs available during inference. Figure~\ref{fig:multi-agent} shows performance comparison between models with and without CECL training across different visual encoders (DINOv2, ViViT, VideoMAE, SigLIP) as the number of selected agents increases from 1 POV to the full team size of 5 POVs.

The results demonstrate a clear pattern: CECL provides substantial performance gains when only 1-2 agent POVs are available. However, as the number of POVs increases toward full team observation (5 agents), the advantage of CECL diminishes. This is expected because when full team observations are available, concatenation-based aggregation already preserves complete information from all viewpoints. In contrast, in low-POV settings, observations are inherently partial and limited, making cross-egocentric alignment more valuable.

CECL's slight underperformance at full-team settings compared to non-contrastive baselines can be attributed to the contrastive objective. While the contrastive objective enhances information sharing among POVs, the training process may lead to a more concentrated embedding space, potentially causing some information loss from the original visual embedding space. This is also indicated by the t-SNE visualization in Figure~\ref{fig:tsne}, where the post-contrastive embeddings form a thinner, more compact manifold. Importantly, CECL is specifically designed for low-POV scenarios—settings in which only a single player's first-person view is available during inference, as would be the case in real gameplay. This could enabling single agent to infer team-level context from its own perspective and paving the way for team-aware AI teammates that reason about collective dynamics from individual experience.

\subsection{Ablation Study}
\label{sec:ablation}

Because the sigmoid loss is sensitive to the large imbalance of positive and negative pairs in a batch, we ablate the values of the counter parameters $b$ and $t$ in the sigmoid loss and validate our design choice of the parameters as mentioned in Section~\ref{sec:cross_egocentric_contrastive_learning}. As shown in Table~\ref{tab:bias_sigmoid}, without proper bias initialization ($b=0$), the model struggles to learn effective representations, while a bias of -3 and temperature of $\log 10$ generally leads to the best performance.

\begin{table}[!ht]
\centering
\caption{Impact of bias term and temperature on contrastive learning performance on DINOv2. The bias $b$ addresses the imbalance between positive and negative pairs in the batch.}
\label{tab:bias_sigmoid}
\begin{tabular}{cccccc}
\toprule
\multirow{2}{*}{\textbf{b}} & \multirow{2}{*}{\textbf{t}} & \multicolumn{2}{c}{\textbf{Teammate}} & \multicolumn{2}{c}{\textbf{Enemy}} \\
\cmidrule(lr){3-4} \cmidrule(lr){5-6}
 &  & \textbf{Sub.Acc} & \textbf{Ham.Dist} & \textbf{Sub.Acc} & \textbf{Ham.Dist}  \\
\midrule
n/a & $\log 10$ & 14.13 & 11.37 & 11.44 & 11.64 \\
0 & $\log 10$ & 13.90 & 11.42 & 9.07 & 11.70 \\
0 & $\log 1$ & 14.68 & 10.51 & 12.15 & 11.36 \\
-3 & $\log 10$ & \textbf{17.01} & \textbf{10.38} & \textbf{12.38} & 11.29 \\
-3 & $\log 1$& 13.47 & 11.17 & 11.68 & 10.98 \\
-10 & $\log 10$ & 15.77 & 10.49 & 10.67 & \textbf{10.96} \\
-10 & $\log 1$ & 1.40 & 12.86 & 1.40 & 12.92 \\
\bottomrule
\end{tabular}
\end{table}

\subsection{Qualitative Evaluation}
\label{sec:qualitative}

\begin{figure*}[ht]
    \centering
    \includegraphics[width=0.95\linewidth]{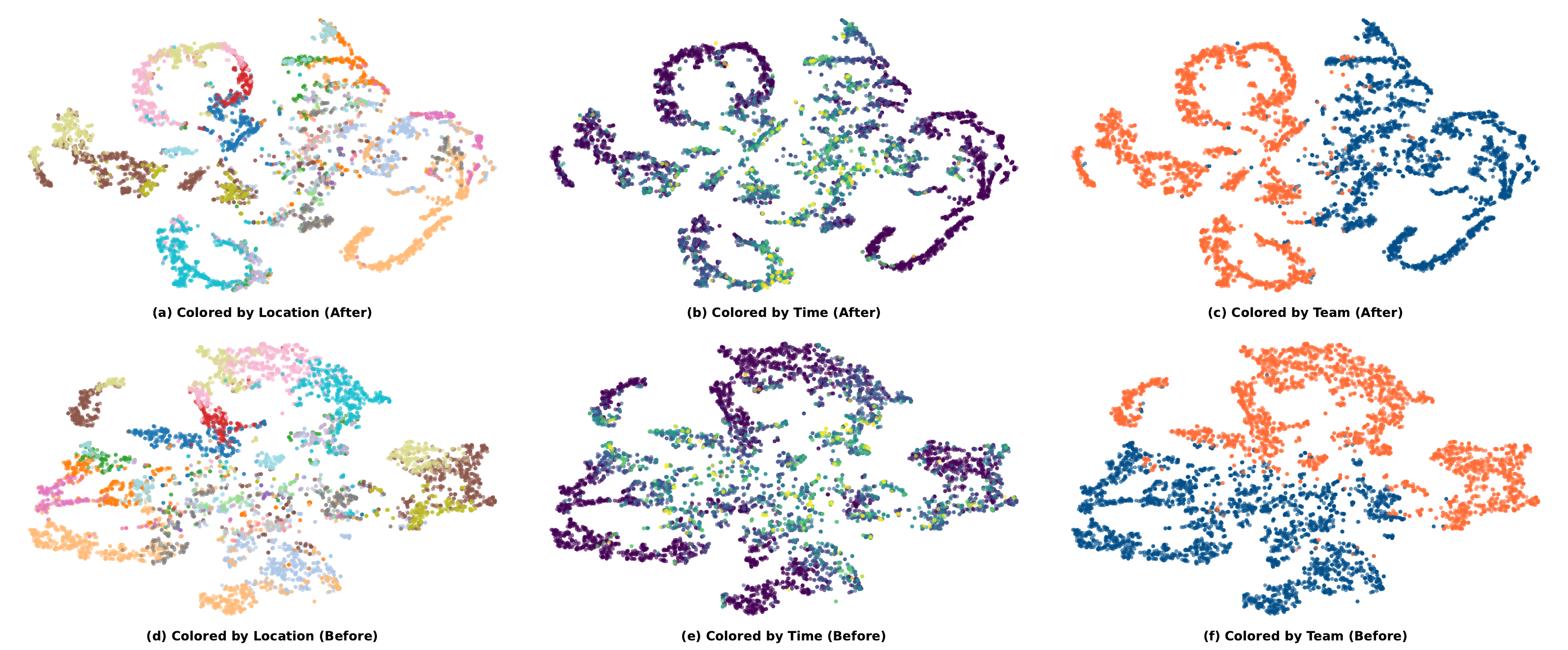}
    \caption{t-SNE visualization of learned embeddings before (bottom row) and after (top row) cross-egocentric contrastive learning. Each point represents a video clip embedding. Columns show different coloring schemes: location (left), time (middle), and team (right). After contrastive training, embeddings demonstrate substantially improved clustering with 8.87\% increase in location-based separation and 15.86\% increase in team-based separation, while maintaining smooth temporal gradients.}
    \label{fig:tsne}
\end{figure*}

We visualize the effectiveness of our cross-egocentric contrastive learning approach using t-SNE dimensionality reduction \cite{maaten2008visualizing}. Figure~\ref{fig:tsne} compares the embedding space before (bottom row) and after (top row) contrastive training, with columns showing different coloring schemes: location (left), time (middle), and team (right). The analysis reveals substantial improvements in representation quality after CECL training, with location-based clustering showing an 8.87\% increase in separation ratio (from 1.272 to 1.385) and team-based clustering demonstrating a 15.86\% improvement (from 1.040 to 1.205). The temporal visualization exhibits clearer cluster margins and smoother gradients after training, indicating that CECL preserves meaningful temporal progression while enhancing spatial clustering.

\section{Discussion}

Our findings suggest that cross-egocentric representation learning offers a promising new paradigm for understanding multi-agent coordination in visually rich, partially observable domains. By aligning teammates' egocentric video streams, CECL enables agents to internalize shared latent representations that approximate a global team state without requiring explicit communication or centralized supervision. This implicit coordination mechanism parallels how human teammates achieve joint situational awareness through mutual observation, anticipation, and inference.

A notable observation is that CECL's benefits are most pronounced under low observability, when only one or two egocentric perspectives are available. This suggests that contrastive alignment can act as an implicit communication channel: by enforcing representational consistency across teammates, the model learns to reconstruct unobserved aspects of the shared environment. In essence, CECL allows an agent to ``imagine'' what its teammates might be perceiving, resembling theory-of-mind reasoning. However, this implicit communication introduces a trade-off as representation compression that can slightly reduce expressivity when full team visibility is available. Future research could explore hybrid contrastive-reconstruction objectives to balance shared and individual encoding fidelity.

Interestingly, CECL did not consistently improve performance on the recent V-JEPA2 0.3B model \citep{assran2025v} in our experimentations, in contrast to other models. One possible explanation is that V-JEPA2's high model capacity and predictive pretraining already encode strong spatiotemporal coherence, leaving limited headroom for additional contrastive regularization. Alternatively, such high capacity models (with over 3 times more parameters than the rest of the models) may require longer training horizons, refined temperature scaling, or customized optimization schedules. 

In addition, our current framework aligns teammates' egocentric representations within the same team and timestep, but does not jointly align all teammates and opponents simultaneously. While this design isolates cooperative information sharing, jointly modeling both sides of the encounter could provide richer tactical structure by capturing inter-team dependencies and adversarial intent. Extending CECL to a fully cross-team alignment setting, where all agents' viewpoints at the same timestep are contrasted within a unified embedding space, represents a promising future direction.

Although our study centers on competitive gaming, the underlying principles extend naturally to other domains involving human-AI teaming under partial observability, such as collaborative robotics, defense operations, and autonomous vehicle fleets. CECL's ability to infer unobserved teammate states from individual sensory inputs points toward scalable models of shared situational awareness in mixed human-machine systems. In these contexts, cross-egocentric alignment could serve as a foundation for interpretable, coordination-aware world models that integrate perception, prediction, and planning.

We identify several promising directions for future work. First, while our experiments focused on contrastive self-supervision for learning team-level awareness, alternative paradigms such as masked video modeling could complement this approach. Inspired by masked language modeling in NLP, future extensions might randomly mask an agent's viewpoint and train a video generator to reconstruct missing perspectives, enhancing cross-agent imagination. Second, cross-egocentric data exist abundantly in real-world human teamwork scenarios—for example, from body-worn or vehicle-mounted cameras—offering opportunities to model multi-human coordination in naturalistic settings. Third, integrating CECL-trained representations into controllable actor agents remains an open direction, enabling direct evaluation of CECL's influence on embodied decision-making and AI-human collaboration.

\section{Conclusion}
We introduced \textbf{X-Ego-CS}, a benchmark dataset for cross-egocentric multi-agent video understanding in professional esports, and proposed \textbf{Cross-Ego Contrastive Learning (CECL)}, a self-supervised framework that aligns teammates' first-person visual streams to acquire team-level situational awareness. Through extensive experiments, we demonstrated that CECL enhances agents' ability to infer both teammate and opponent positions from limited egocentric views, particularly under partial observability, highlighting the potential of cross-egocentric alignment as a scalable mechanism for fostering implicit coordination and shared tactical understanding without explicit communication or supervision. More broadly, this work establishes a foundation for collective perception and reasoning in multi-agent systems, bridging self-supervised representation learning and human-AI teaming, and opens new pathways for understanding and reproducing team dynamics in complex, real-time environments through cross-perspective alignment.

\begin{acks}
The project or effort depicted was or is sponsored by the U.S. Army Combat Capabilities Development Command -- Soldier Centers under contract number W912CG-24-D-0001. The content of the information does not necessarily reflect the position or the policy of the Government, and no official endorsement should be inferred. The authors acknowledge the use of Large Language Models for assistance with proofreading and grammar checking. All content was reviewed, edited, and approved by the human authors, who take full responsibility for the final manuscript. 
\end{acks}

\bibliographystyle{ACM-Reference-Format} 
\bibliography{ref}


\end{document}